\documentclass[]{spie}  

\usepackage{amsmath,amsfonts,amssymb}
\usepackage{graphicx}
\usepackage[colorlinks=true, allcolors=blue]{hyperref}
\usepackage[acronym]{glossaries}
\usepackage{bbold} 

\usepackage[normalem]{ulem}
\useunder{\uline}{\ul}{}

\usepackage[T1]{fontenc}
\usepackage{currvita}

\usepackage{booktabs, multirow} 

\usepackage{siunitx}

\newacronym{oct}{OCT}{optical coherence tomography}
\newacronym{cnn}{CNN}{convolutional neural network}
\newacronym{tv}{TV}{total variation}

\title{Attenuation artifact detection and severity classification in intracoronary OCT using mixed image representations}

\author[a,b,c]{Pierandrea Cancian}
\author[a,b,c]{Simone Saitta}
\author[a,b,c]{Xiaojin Gu}
\author[a,b,c]{Rudolf L.M. van Herten}
\author[d,e]{Thijs J. Luttikholt}
\author[d,e]{Jos Thannhauser}
\author[d]{Rick H.J.A. Volleberg}
\author[d,e]{Ruben G.A. van der Waerden}
\author[d,e]{Joske L. van der Zande}
\author[b,a]{Clarisa I. Sánchez}
\author[d]{Bram van Ginneken}
\author[e]{Niels van Royen}
\author[a,b,c,f]{Ivana Išgum}
\affil[a]{Quantitative Healthcare Analysis group, Biomedical Engineering and Physics, Amsterdam~UMC, the Netherlands.}
\affil[b]{Quantitative Healthcare Analysis group, Informatics Institute, University~of~Amsterdam,~the~Netherlands}
\affil[c]{Amsterdam Cardiovascular Sciences, Amsterdam UMC, the Netherlands.}
\affil[d]{Diagnostic Image Analysis Group, Radboud University Medical Center, Nijmegen,~the~Netherlands.}
\affil[e]{Department of Cardiology, Radboud University Medical Center, Nijmegen,~the~Netherlands.}   
\affil[f]{Department of Radiology and Nuclear Medicine, Amsterdam UMC, the Netherlands.}   

\authorinfo{Further author information: (Send correspondence to P.C.)\\P.C.: E-mail: p.cancian@amsterdamumc.nl}

\begin{document} 
\maketitle

\begin{abstract}
In intracoronary optical coherence tomography (OCT), blood residues and gas bubbles cause attenuation artifacts that can obscure critical vessel structures. The presence and severity of these artifacts may warrant re-acquisition, prolonging procedure time and increasing use of contrast agent. Accurate detection of these artifacts can guide targeted re-acquisition, reducing the amount of repeated scans needed to achieve diagnostically viable images.
However, the highly heterogeneous appearance of these artifacts poses a challenge for the automated detection of the affected image regions.
To enable automatic detection of the attenuation artifacts caused by blood residues and gas bubbles based on their severity, we propose a convolutional neural network that performs classification of the attenuation lines (A-lines) into three classes: no artifact, mild artifact and severe artifact. 
Our model extracts and merges features from OCT images in both Cartesian and polar coordinates, where each column of the image represents an A-line. 
Our method detects the presence of attenuation artifacts in OCT frames reaching F-scores of $0.77$ and $0.94$ for mild and severe artifacts, respectively. The inference time over a full OCT scan is approximately 6 seconds.
Our experiments show that analysis of images represented in both Cartesian and polar coordinate systems outperforms the analysis in polar coordinates only, suggesting that these representations contain complementary features. 
This work lays the foundation for automated artifact assessment and image acquisition guidance in intracoronary OCT imaging.
\end{abstract}


\section{INTRODUCTION}
\label{sec:intro}  
\begin{figure}[t]
  \begin{center}
  \begin{tabular}{c} 
  \includegraphics[height=2.73cm]{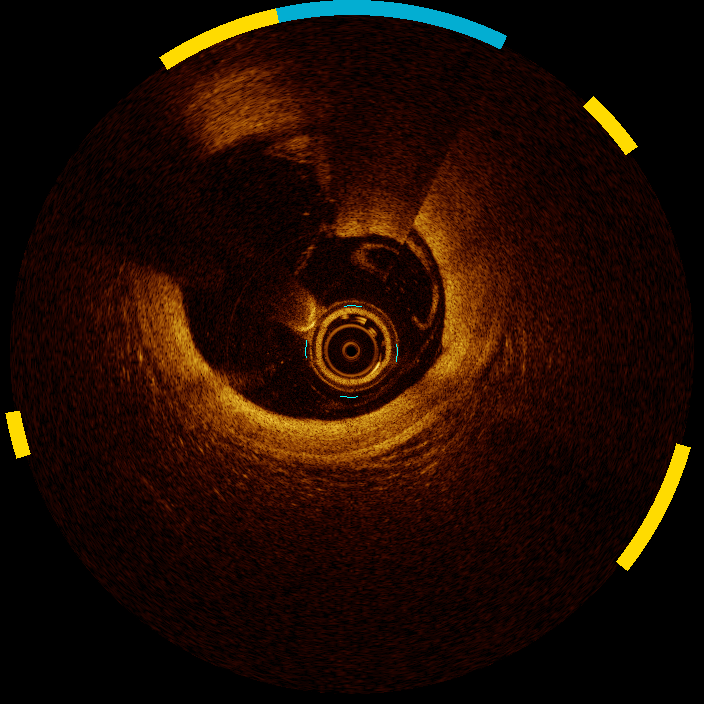}
  \includegraphics[height=2.73cm]{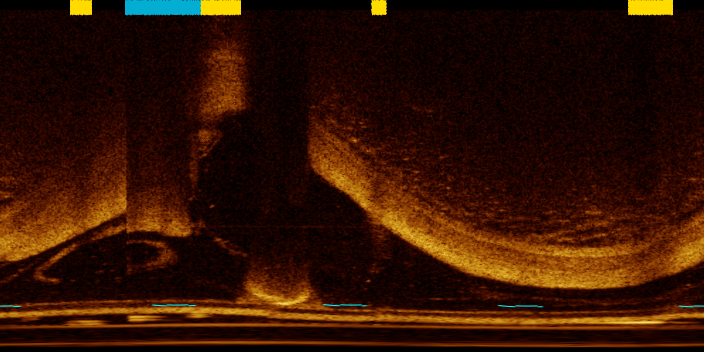}
  \includegraphics[height=2.73cm]{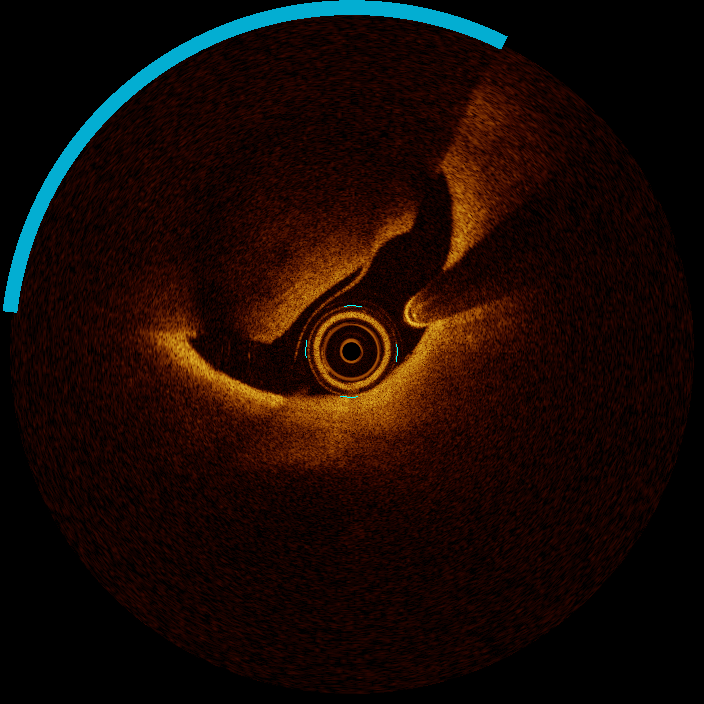}
  \includegraphics[height=2.73cm]{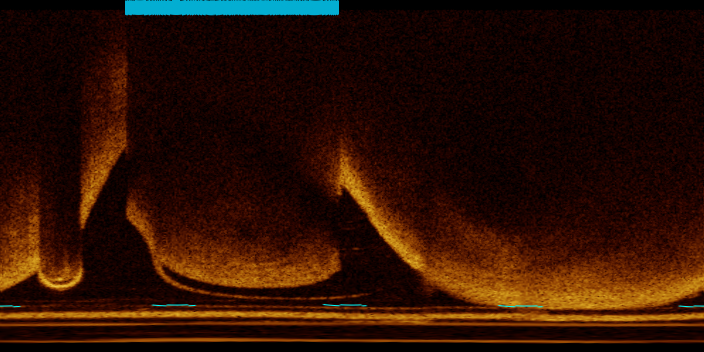}
  \end{tabular}
  \end{center}
  \caption[example] 
  {\label{fig:data}
  Examples of two optical coherence tomography images in the Cartesian domain and their polar counterpart, with annotated artifacts. The first and third image from the left show frames in Cartesian coordinates and the second and fourth in polar coordinates. Mild artifacts are indicated in yellow and severe in blue. Arcs in the Cartesian domain correspond to straight segments in polar coordinates, where the columns of the image are equivalent to the original A-lines. In polar coordinates the artifacts and their shadows are vertically aligned.}
\end{figure}

Intracoronary \gls{oct} is an invasive imaging modality that yields consecutive high-resolution cross-sectional images of a coronary artery segment. OCT can offer accurate information on the vessel microstructure, such as plaque presence and characteristics, as well as details regarding the state of stented lesions\cite{araki2022optical}.
This imaging technique has been proven to provide valuable complementary information to X-ray angiography in guiding percutaneous revascularization~\cite{holm2023oct,stone2024intravascular}. 
During intracoronary \gls{oct} acquisition, the OCT probe emits near-infrared light in a helical pattern, acquiring multiple attenuation lines (A-lines) and forming a 3D volume, which is referred to as a \textit{pullback}. 
An OCT pullback consists of a few hundred cross-sectional images of the coronary artery in Cartesian coordinates, referred to as \textit{frames}. 
Each frame typically has an in-plane resolution of 10-15 \textmu m, and the length of the pullback can reach 150 mm.
During \gls{oct} imaging, blood must be completely flushed from within the lumen. Incomplete flushing can leave blood residues in the lumen and cause the formation of gas bubbles within the catheter. Blood residues and gas bubbles cause attenuation artifacts, casting shadows that prevent the light from reaching the arterial wall (Figure \ref{fig:data}).  
The presence of these artifacts may mask critical features, leading to incorrect assessment of an intracoronary \gls{oct} exam and resulting in misdiagnosis or suboptimal decision making.
Automated detection and severity assessment of these artifacts could provide important feedback to the operator; the detection of severe attenuation artifacts can suggest re-acquisition of the pullback, whereas detection of mild artifacts can provide useful information during image interpretation and help identify areas where minor corrections can enhance the overall image quality.

Despite the relevance of automated image quality assessment, to the best of our knowledge no method on attenuation artifacts detection in intracoronary \gls{oct} has been presented. 
Rather, earlier studies focused on the correction of a specific class of artifacts, most commonly the shadow cast by the OCT guidewire. 
For example, Gharaibeh et al.~\cite{gharaibeh} approached guidewire artifact and stent struts' shadow inpainting with conditional generative adversarial networks.
Thereafter, Luo et al.\cite{trajectoryaware} devised a sophisticated deep learning solution for guidewire inpainting that leveraged both texture-focused and structure-focused frames. 
Tao et al.\cite{Defocus} developed a physics-inspired model to correct contrast agent attenuation and OCT defocus. 
Jessney et al.~\cite{JESSNEY2024} showed how the correction of common mild artifacts, such as gas bubbles, blood residues, and seam artifacts, through a Fourier transform-based approach can improve identification of higher-risk features and plaque classification.
Instead of focusing on artifact correction, Cheimariotis et al.\cite{Cheimariotis} developed a multi-step image processing pipeline for lumen border detection that is resistant to blood residues. 
These studies underscore the importance of dealing with artifacts to improve \gls{oct} image analysis. 

The objective of the present study is to develop an automatic method for the detection of attenuation artifacts caused by blood residues or gas bubbles according to their severity. 
For this purpose, we propose a \gls{cnn} that performs A-line classification into severity classes, exploiting the inherent radial structure of the attenuation artifacts.

\section{DATA}
We included 80 pullbacks from 75 patients from PECTUS-obs\cite{mol2023fractional,mol2021identification}, a prospective observational study that includes pullbacks with non-flow-limiting, nonculprit lesions of patients with myocardial infarction.
All patients provided informed consent. 

The reference standard included three classes indicating artifact presence and severity: \textit{none}, \textit{mild} and \textit{severe}. 
The reference classes did not distinguish whether the attenuation was due to a gas bubble or a blood artifact. 
Mild artifacts were defined as those casting faint shadows that still allowed discerning the underlying structures, while severe artifacts made interpretation impossible.

To define the reference standard, A-lines were annotated using the brush tool within ITK-SNAP\cite{itksnap} to paint the corresponding label over the artifact-affected arcs. 
This operation was performed on images visualized in Cartesian coordinates. Figure \ref{fig:data} shows an example of the annotated arcs.
The annotation of each frame was then processed to derive a vector $\mathbf{y} \in \{0,1,2\}^{\vartheta}$, where $\vartheta$ is the number of A-lines per frame.
The pullbacks and frames selected for annotation were chosen such that the annotated dataset would be heterogeneous in terms of artifact appearance. 
Possible variations in appearance included high versus low mixing of blood residues within the lumen and the size of the gas bubbles.
The dataset also included frames featuring red thrombi, pathological structures that visually resemble blood artifacts.

\begin{figure}[b]
  \begin{center}
  \begin{tabular}{c} 
      \includegraphics[width=.97\linewidth]{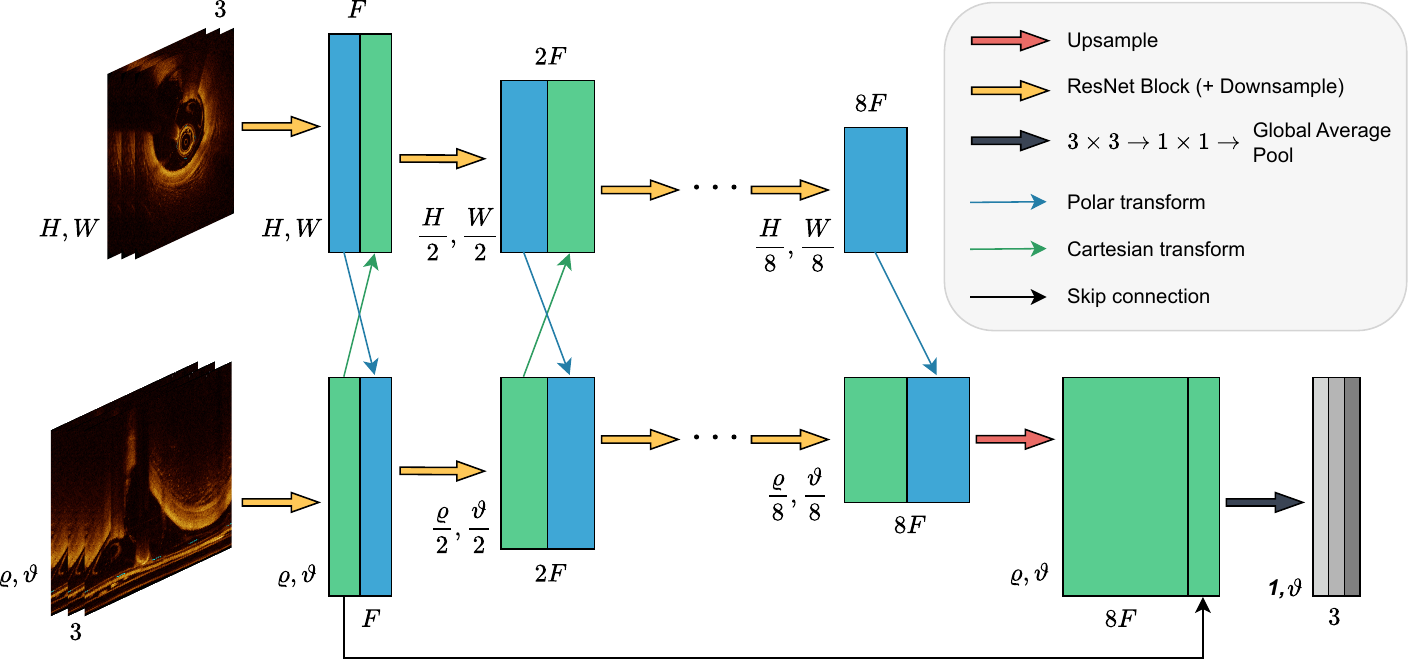}
  \end{tabular}
  \end{center}
  \caption[example] 
  {\label{fig:anet_diagram}
  Schematic of the ArcNet model. The top branch takes as input the Cartesian image, while the polar transformed image is fed to the bottom branch. Blocks represent feature dimensions (blue for Cartesian features and green for polar). $[\, H,W\, ]$ and $[\, \varrho, \vartheta\, ]$ are image sizes, F is feature size and it refers to the output of the last convolution before the concatenation with the features from the other domain. Output has size $[\, \vartheta, 3\, ]$, where each element in $\vartheta$ represents an A-line and 3 represents the number of classes.}
\end{figure}

\section{METHODS}
\subsection{Model architecture}
We propose a \gls{cnn}, referred to as ArcNet, that processes intracoronary \gls{oct} images to identify A-lines affected by attenuation artifacts. 
It does so by classifying each A-line into one of three artifact classes: \textit{none}, \textit{mild} and \textit{severe}. 
The method employs a \gls{cnn} that leverages \gls{oct} images in both Cartesian and polar coordinate systems.
The former, depicts the undistorted anatomical view routinely used for image interpretation. 
The latter, transforms the A-lines from radial lines originating in the center of the image into the columns of the polar image.
In our work, the polar transform is always performed using the center of the image as the reference point for the transform, which is always the center of the acquisition catheter.
To obtain the pixel values in the polar transform, or its inverse, we employ bilinear interpolation. 

The proposed \gls{cnn} architecture features two dedicated branches, one for the polar and one for the Cartesian images, that exchange information after each computation block to create a mixed representation (Figure \ref{fig:anet_diagram}).
Both branches take as input three consecutive grayscale images, stacked in the channel dimension to provide through-plane context. 
The Cartesian branch accepts a stack of Cartesian images of size 
$[\, 3 \! \times \! H \! \times \! W \,]$ with 
$H \! = \! W \! = \! 352$ corresponding to half the dimensions of the imaging system's output, while the polar branch takes as input the polar-transformed counterpart of size 
$[\, 3 \times \varrho \times \vartheta \,]$,
where 
$\varrho \!= \! 176$ and $\vartheta \! = \! 224$
are respectively the radial and angular dimensions. 

The polar branch consists of 4 residual blocks~\cite{he2015deepresiduallearningimage}, modified to use circular padding in the $\vartheta$ dimension, with feature sizes $[\,32, 64, 128, 256\,]$ and downsampling factors $[\,2, 2, 2, 1\,]$ with strided convolutions.
The Cartesian branch has the same configuration, with half the feature sizes and standard zero padding.
The information exchange between the two branches occurs by concatenating the features of one branch with those of the other branch before feeding them into the next computational block. 
To maintain spatial coherence, the polar transform is applied to the feature volume coming from the Cartesian branch prior to being concatenated to the feature volume from the polar branch. 
Similarly, for features moving in the opposite direction, the inverse polar transform is used.
After the last block of the polar branch, the feature volume is upsampled to match the shape of the output of the first block and then concatenated with them to recover finer local details. The feature volume is projected to the number of classes with $1 \times 1$ convolutions and the $\varrho$ dimension is collapsed through global average pooling to yield a vector $\mathbf{x} \in \mathbb{R}^{\vartheta \times 3}$ where each $\mathbf{x}_i \in \mathbb{R}^3$ represents the prediction of the $i$-th A-line in one of the three classes (\textit{none}, \textit{mild} and \textit{severe}). 

\subsection{Loss function}
From the output logits $\mathbf{x}$ we compute $p_{i,c} = \sigma(\mathbf{x}_i)_c$ which is the probability of class $c$ at location $i$ computed with the softmax operator. 
The loss function is then computed as the mean of the cross-entropy loss and the soft 1D Dice loss with the addition of a small regularization term based on the \gls{tv} of the logits.
Let $\mathbf{y} \in \{0, 1, 2\}^\vartheta$ be the reference annotation,
the formula for the soft Dice loss on a single output is as follows:
$$
\mathcal{L}_{\text{Dice}}(\mathbf{x}, \mathbf{y}) = 1 
     - \frac{1}{C}\sum_{c=0}^{C-1} \sum_{i=1}^{\vartheta} \, 
     \frac{2 \, p_{i,c} \: \mathbb{1} \{c=y_{i}\}}
          {\: 2 \, p_{i,c} \: \mathbb{1} \{c=y_{i}\} 
           + p_{i,c} \: \mathbb{1} \{c\neq y_{i}\}
           + (1-p_{i,c}) \:\mathbb{1} \{c=y_{i}\} \:},
$$
where $\mathbb{1} \{condition\}$ is the indicator function assuming value $1$ when the \textit{condition} is met and $0$ otherwise. 
While the \gls{tv} over a single sample is computed independently for each class and then averaged:
$$
\mathcal{L}_{TV} = \frac{1}{C}\sum^{C-1}_{c=0} \| \nabla \mathbf{x}_c \|_1.
$$
To compute the gradient at the edges, we apply circular padding to $\mathbf{x}$ in the $\vartheta$-dimension. 
The formula for the complete loss function then becomes:
$
\mathcal{L} = \frac{1}{2}\mathcal{L}_{\text{CE}} + \frac{1}{2} \mathcal{L}_{\text{Dice}} + \lambda \: \mathcal{L}_{TV}
$,
with $\lambda = 5\cdot10^{-4}$ being a scaling coefficient.

\subsection{Data sampling}
OCT pullbacks typically contain a limited number of frames with artifacts, and these artifacts affect only a portion of the A-lines. 
To account for the typical over-representation of frames without artifacts and the significant variability in the number of artifacts within individual frames, we designed a stratified sampling strategy to ensure the model was exposed to a diverse set of samples during training.
To increase the likelihood of drawing under-represented samples, we regulate the probability of drawing each sample in the training set through a proxy metric that estimates the number of A-lines with artifacts. 
Let $Y = [\mathbf{y}_1, ..., \mathbf{y}_{N_t}]$ be the list of all reference annotations in the training set. For the n-$th$ annotated frame, we compute the proxy metric $$K_n = \lceil log(1 + \sum_\vartheta \mathbf{y}_n) \rceil,$$ hence, forming an array $K = [K_1, ..., K_{N_t}]$ for the entire dataset. Let $M$ be the cardinality of the most frequent value in $K$, and $|K_n|$ the number of occurrences of the n-$th$ element of $K$. Then, the sampling weight of the n-$th$ frame is computed as: 
$$w_n = \min(\frac{M}{|K_n|}, 50).$$
The clipping ensures that no sample is drawn \textit{too} often, which could become counter-productive by leading to over-fitting.

\section{Evaluation}
To assess the performance of the presented method at different levels of granularity, a range of metrics were computed.

First, the classification performance on A-lines was evaluated on the test set. For each frame, the 1D Dice score was computed to measure the overlap between model predictions and reference annotations. These scores were then averaged across all test set frames and are presented alongside their standard deviations. This metric was only calculated for frames where both the predicted and ground truth classes were present.
To account for class imbalance, balanced accuracy was computed as the average of recall obtained for each class, ensuring equal class contribution to the overall performance metric.
Additionally, A-line F-scores were computed over the whole A-line dataset, irrespective of the image they came from. 
Furthermore, confusion matrices were generated, providing a percentage breakdown of predictions for each class and offering comprehensive insights into the accuracy and errors of the model.

Next, frame-level F-scores were computed to evaluate the model's ability to detect frames containing specific artifact classes. This metric allows to evaluate whether a given class was correctly predicted in a frame. Unlike A-line metrics, frame-wise metrics are binary (presence/absence) assessments, providing insights into the model's performance in detecting artifact-affected frames rather than the segmentation quality within each frame.

\section{Experiments and results}
\label{sec:results}
In total, 7,543 annotated frames were divided into training, validation, and test sets at a 70:15:15 ratio at the patient level. 
Among these, 4,660 frames did not contain any artifact, while mild and severe artifacts were present in 2,173 and 1,611 frames, respectively. 
In the training set, 30\% of the frames contained mild artifacts and 19\% contained severe artifacts. In the validation set these percentages were 27\% and 26\%; while in the test set, mild and severe artifacts were present in 24\% and 33\% of frames, respectively.

The ArcNet was trained with the Adam\cite{kingma2017adam} optimizer, with a learning rate of $10^{-5}$ reduced by a factor of two after five epochs without improvement on the validation loss. 
The training lasted 100 epochs, each including 250 batches of size 12.
The model with the lowest validation loss was selected to be evaluated on the test set.
Data augmentation included random rotations, horizontal flipping and brightness changes. 

\begin{figure}[h]
  \begin{center}
  \includegraphics[width=\textwidth]{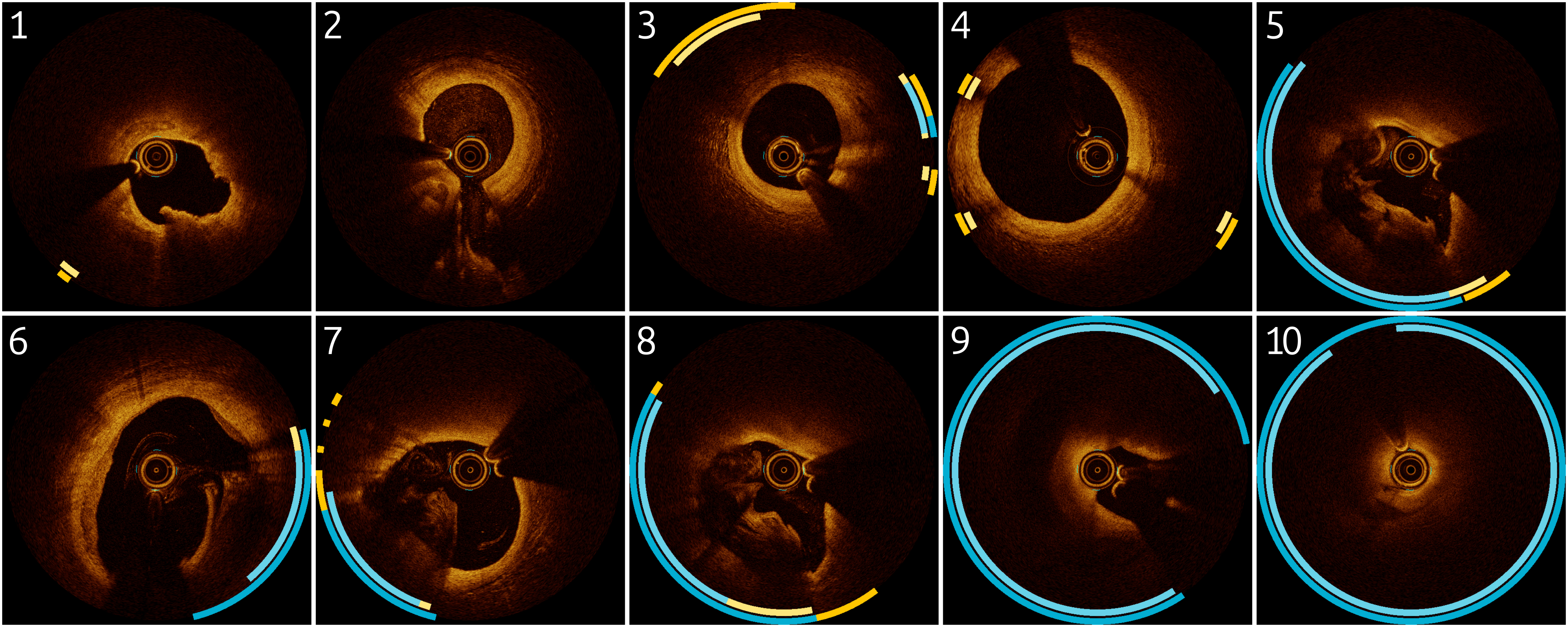}
  \end{center}
  \caption[example] 
  {\label{fig:predictions}
  Example predictions of the ArcNet. Mild artifacts are indicated in yellow and severe in blue. External ring shows reference annotations and internal shows the prediction of our method. 1) Red thrombus does not trigger false positives, 2) Complex sample with no artifact, but presenting sidebranches, plaques, highly-mixed non-attenuating blood and tangential signal drop-off, 3-4) Mild attenuation due to gas bubbles, 5-6) Severe blood artifacts, 7) Small mild artifacts at 10 o'clock are not predicted, likely due to low input image resolution, 8) Imprecise segmentation of part of the severe artifacts 9-10) Near-perfect segmentation of very severe blood artifacts.}
\end{figure}

Quantitative results of the proposed ArcNet are listed in Table \ref{tab:results}, top row.
Figure \ref{fig:predictions} illustrates its predictions, demonstrating accuracy for a wide range of artifact appearances and anatomical variations. The ArcNet, processed a full pullback of about 500 frames in approximately 6 seconds on a NVIDIA GeForce RTX 3080 Ti with 12 GB of memory.

\begin{table}[h]
\caption{
\label{tab:results}
The best performance for each metric is shown in \textbf{bold} and {\ul underlined} in case of a tie. \\From the second to the sixth column, A-line metrics are listed. In the seventh and eighth columns, frame-wise metrics are presented.}
\vspace{4pt}
\centering
\small
\begin{tabular}{|l|c|c|c|c|c|c|c|}
\hline
\multicolumn{1}{|c|}{\rule[-1ex]{0pt}{3.5ex}Model name} &
  \begin{tabular}[c]{@{}c@{}}Balanced\\Accuracy\end{tabular} &
  \begin{tabular}[c]{@{}c@{}}1D Dice\\Mild\end{tabular} &
  \begin{tabular}[c]{@{}c@{}}1D Dice\\Severe\end{tabular} &
  \begin{tabular}[c]{@{}c@{}}F-score\\Mild\end{tabular} &
  \begin{tabular}[c]{@{}c@{}}F-score\\Severe\end{tabular} &
  \begin{tabular}[c]{@{}c@{}}Frame-wise\\F-score  Mild\end{tabular} &
  \begin{tabular}[c]{@{}c@{}}Frame-wise\\F-score Severe\end{tabular} \\ \hline
\rule[-1ex]{0pt}{3.5ex}ArcNet & {\ul 0.71} & 0.43 $\pm$ 0.29  & 0.80 $\pm$ 0.20  & 0.39  & 0.85  & \textbf{0.77} & \textbf{0.94} \\ \hline
\rule[-1ex]{0pt}{3.5ex}ArcNet One-Way & 0.70  & 0.40 $\pm$ 0.27  & 0.80 $\pm$ 0.21   & 0.31  & 0.86  & 0.63  & 0.91  \\ \hline
\rule[-1ex]{0pt}{3.5ex}ArcNet Single & {\ul 0.71} & 0.41 $\pm$ 0.28  & \textbf{0.83 $\pm$ 0.18} & 0.32  & \textbf{0.87} & 0.61  & 0.92  \\ \hline
\rule[-1ex]{0pt}{3.5ex}ArcNet Polar  & 0.67  & 0.38 $\pm$ 0.27  & 0.78 $\pm$ 0.21  & 0.25  & 0.84  & 0.58  & 0.92  \\ \hline
\rule[-1ex]{0pt}{3.5ex}FanCNN* 2D  & 0.67  & 0.40 $\pm$ 0.27  & 0.75 $\pm$ 0.22  & 0.39  & 0.71  & 0.67  & 0.78  \\ \hline
\rule[-1ex]{0pt}{3.5ex}FanCNN* 3D  & {\ul 0.71} & \textbf{0.44 $\pm$ 0.30} & 0.77 $\pm$ 0.23  & \textbf{0.40} & 0.77  & 0.65  & 0.88  \\ \hline
\end{tabular}
\end{table}

To evaluate the impact of different levels of information exchange between the Cartesian and polar branches, we compared ArcNet with variants of its architecture that feature fewer connections between the parallel branches. 
While ArcNet features a bi-directional exchange of information, the ArcNet One-Way exchanges information unidirectionally from the Cartesian branch to the polar branch. 
The ArcNet Single only injects the output of a single Cartesian block to the polar branch at the beginning of the backbone. Finally, the ArcNet Polar has no Cartesian branch at all. All ArcNet variants were trained following the same procedure described for the original ArcNet model. 
The results of these three variants are listed in Table~\ref{tab:results}, second to forth rows.

Among the tested configurations, the highest frame-wise F-scores were achieved by the ArcNet itself, 0.77 for mild and 0.94 for severe artifacts. In particular, for the mild class, the ArcNet outperformed by 0.19 the ArcNet Polar, the worst performing model. 
Regarding the F-scores at the A-line level, the performances for the severe artifact class were comparable across all tested settings, while for the mild class the ArcNet Polar showed inferior performance (F-score 0.25) compared with the other variants featuring mixed polar and Cartesian representations. 
Here, the ArcNet (F-score 0.39) outperformed the ArcNet One-Way (F-score 0.31) and the ArcNet Single (F-score 0.32), which only have partial exchange of information between the two branches.
All tested settings achieved similar performance in terms of 1D Dice score and balanced accuracy.

\begin{figure}[t]
  \begin{center}
  \includegraphics[width=\textwidth]{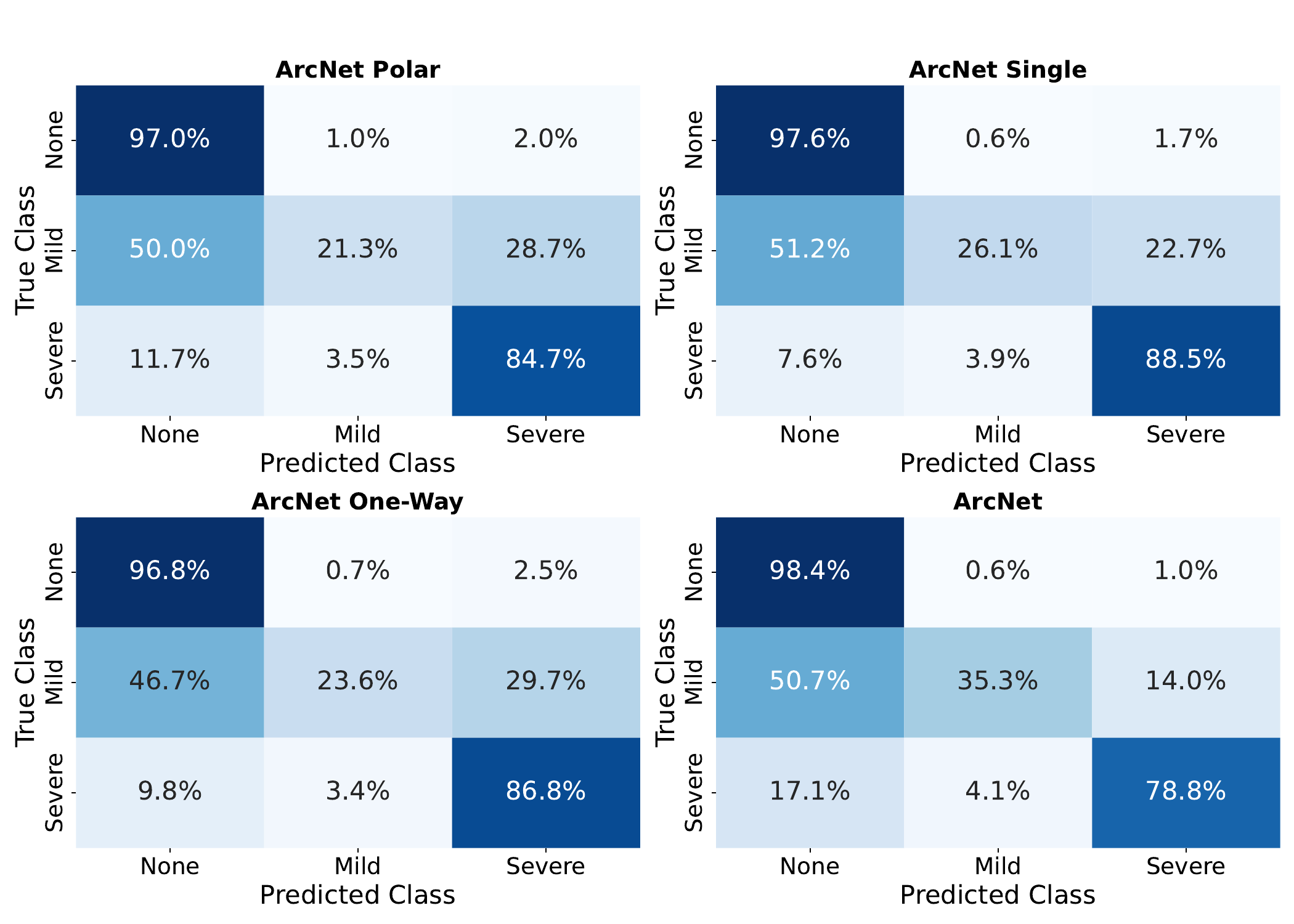}
  \end{center}
  \caption[example] 
  {\label{fig:confusion_mats}
  Confusion matrices of the four ArcNet models (ArcNet Polar, ArcNet Single, ArcNet One-Way, and ArcNet) showing the predicted class percentages for each true class (\textit{none}, \textit{mild}, \textit{severe}) at the A-line level. The matrices are normalized to display percentages.}
\end{figure}

The confusion matrices in Figure \ref{fig:confusion_mats} provide further insights into the classification performance at the A-line level of the ArcNet and its variants. 
All the tested settings misclassified a significant portion of \textit{mild} as the \textit{none} class (up to 51.2\% for the ArcNet Single) and in some cases confused mild artifacts with severe ones (up to 29.7\% for the ArcNet One-Way). ArcNet demonstrated superior performance with respect to its variants in accurately classifying mild artifacts, with only 14.0\% of mild artifacts being misclassified as severe. 
Nonetheless, ArcNet missed more severe artifacts (17.1\%) than the other variants, with the ArcNet Single misclassifying only 7.6\% of the severe artifacts into the \textit{none} class.

Additionally, we performed a comparison with another model developed to work in polar coordinates, FanCNN \cite{vanherten2023automatic}. For application to coronary OCT, we adapted the FanCNN originally created for the analysis of multi-planar reformatted coronary arteries from computed tomography angiography. 
For this, we increased the number of output channels in the last fully connected layer to accommodate the higher resolution of \gls{oct} images and changed the number of heads to match the number of artifact classes.
We trained and evaluated both the original 3D network (FanCNN* 3D), taking as input 15 adjacent OCT frames, and a 2D counterpart (FanCNN* 2D), operating on individual \gls{oct} frames. FanCNN* models were trained using the same training procedure described for the ArcNet models.

For severe artifacts, ArcNet and its variants achieved higher mean 1D Dice and F-scores compared to the FanCNN* baselines, both at the A-line level and the frame level. 
For A-lines affected by mild artifacts, similar performances were observed among the ArcNet models and the FanCNNs*, with the FanCNN* 3D achieving the highest 1D Dice and F-score, on average. 

\section{DISCUSSION}
We presented a method for detecting attenuation artifacts in intracoronary \gls{oct} by classifying A-lines into severity classes. Framing the detection problem as A-line classification represents a trade-off between frame-level classification and pixel-wise segmentation. 
Frame classification is unsuitable for attenuation artifact detection, as it does not provide feedback on which parts of the frame are affected by artifacts. 
Additionally, it would require setting a subjective threshold to determine which frames are too corrupted for interpretation. 
Conversely, full pixel-wise semantic segmentation of shadow-casting artifacts would introduce unnecessary redundancy, as precise delineation is not needed since these artifacts do not overlap with the structures they affect. A-line classification strikes a balance by enabling fine-grained identification of both the artifacts and the regions they obscure, while simplifying model design and annotation.

In our study, the proposed architecture reached the highest performance at the frame level, while remaining competitive on all the other metrics, suggesting that the bi-directional flow of information between the two branches allows the extraction of more effective features. 
The polar branch exploits the alignment of artifacts with the shadows they cast, promoting precise A-line classification, whereas the Cartesian branch captures the undistorted vessel anatomy, where artifacts exhibit recognizable patterns.
This dual-branch architecture is particularly advantageous for clinical applications, as frame-level detection provides rapid feedback on pullback quality, which is critical for real-time decision-making during procedures.

We observed that the ArcNet makes fewer spurious small predictions of mild artifacts compared to all other models, likely due to the more discriminative features extracted from the reciprocal information exchange of its two branches. 
The performance difference between the ArcNet models and the FannCNN* models is likely caused by the difference in data resolution: OCT data has higher resolution than the computed tomography data the FanCNN was designed for. Introducing skip connections and downsampling blocks potentially enables the ArcNet models to learn more relevant hierarchical features with respect to the FanCNN*.

\Gls{tv} regularization was included in the loss function of all our models. In our preliminary experiments, we found that adding \gls{tv} regularization increased the smoothness of the predictions, making them less fragmented. 

Significant limitations of our study include the lack of standardization in the number of trainable parameters across different models and the use of a limited dataset for evaluation. 
Furthermore, performance in detecting mild artifacts remains lower compared to severe artifacts. 
Mild artifacts often exhibit more nuanced appearances, are smaller in size, and have subjective and inconsistent boundaries that make classification challenging. 
Further work focus on gathering more annotated data to ensure greater diversity in the training set. Experimenting with class weights in the loss function could offset the consistently smaller size of mild artifacts with respect to severe artifacts.

\section{Conclusion}
We presented a method to detect attenuation artifacts through the classification of their severity in intracoronary \gls{oct}, showcasing the benefits of using mixed Cartesian and polar feature representations for A-line classification. Our approach lays the foundations for automated and accurate artifact assessment in intracoronary \gls{oct} acquisitions.

\acknowledgments
This publication is part of the project ROBUST: Trustworthy AI-based Systems for Sustainable Growth with project number KICH3.LTP.20.006, which is (partly) financed by the Dutch Research Council (NWO), Abbott, and the Dutch Ministry of Economic Affairs and Climate Policy (EZK) under the program LTP KIC 2020-2023.

\bibliography{report} 
\bibliographystyle{spiebib} 

\end{document}